\documentclass[sigconf]{aamas} 

\usepackage{balance} 
\usepackage{algorithm}
\usepackage{algorithmic}
\usepackage{amsmath}
\usepackage{newfloat}
\usepackage{listings}

\setcopyright{ifaamas}
\acmConference[AAMAS '22]{Proc.\@ of the 21st International Conference
on Autonomous Agents and Multiagent Systems (AAMAS 2022)}{May 9--13, 2022}
{Auckland, New Zealand}{P.~Faliszewski, V.~Mascardi, C.~Pelachaud,
M.E.~Taylor (eds.)}
\copyrightyear{2022}
\acmYear{2022}
\acmDOI{}
\acmPrice{}
\acmISBN{}

\acmSubmissionID{???}

\title{Learning General Inventory Management Policy\\ for Large Supply Chain Network}

\author{Soh Kumabe}
\affiliation{
  \institution{The University of Tokyo}
  \city{Tokyo}
  \country{Japan}}
\email{soh_kumabe@mist.i.u-tokyo.ac.jp}

\author{Shinya Shiroshita}
\affiliation{
  \institution{Preferred Networks Inc.}
  \state{Tokyo}
  \country{Japan}}
\email{shiroshita@preferred.jp}

\author{Takanori Hayashi}
\affiliation{
  \institution{Preferred Networks Inc.}
  \state{Tokyo}
  \country{Japan}}
\email{thayashi@preferred.jp}

\author{Shirou Maruyama}
\affiliation{
  \institution{Preferred Networks Inc.}
  \state{Tokyo}
  \country{Japan}}
\email{maruyama@preferred.jp}

\begin{abstract}
Inventory management in warehouses directly affects profits made by manufacturers. Particularly, large manufacturers produce a very large variety of products that are handled by a significantly large number of retailers. In such a case, the computational complexity of classical inventory management algorithms is inordinately large.
In recent years, learning-based approaches have become popular for addressing such problems. However,  previous studies have not been managed systems where both the number of products and retailers are large.
This study proposes a reinforcement learning-based warehouse inventory management algorithm that can be used for supply chain systems where both the number of products and retailers are large. To solve the computational problem of handling large systems, we provide a means of approximate simulation of the system in the training phase. Our experiments on both real and artificial data demonstrate that our algorithm with approximated simulation can successfully handle large supply chain networks.\footnote{This result is based on the first author's internship program at Preferred Networks. Inc.}
\end{abstract}


\newcommand{\BibTeX}{\rm B\kern-.05em{\sc i\kern-.025em b}\kern-.08em\TeX}

\begin{document}


\pagestyle{fancy}
\fancyhead{}


\maketitle 

\section{Introduction}

\subsection{Background and Motivation}

The supply chain networks of major manufacturers are huge. For example, Nestle sells over 2,000 brands of products to its retailers worldwide. Because of the enormous size of these networks, even a small improvement in delivery planning can significantly reduce the costs of a company.

For this reason, inventory management has been studied for a long time. In practice, a supply chain network is modeled by several facilities (factories, warehouses, and retailers) and the delivery routes between them. Each facility delivers products only to facilities further downstream -- factories to warehouses or retailers, or warehouses to other warehouses or retailers. The goal is to maximize the gain of the network by deciding which products should be delivered, when they should be delivered, and the quantities to deliver along each delivery route.

The difficulty in inventory management lies in managing the amount of inventory in a facility. If a warehouse has very little inventory, it will run out of products frequently or even permanently. It will often not be able to meet the retailers’ requests, which can severely affect profits.
In contrast, having a large inventory in a warehouse is also a problem. Inventory occupies space in the warehouse. In the case of products with expiration dates, excess inventory leads to disposal. The value of a product with an expiration date will diminish over time. Therefore, excess inventory can be regarded as a cost in itself. Balancing the trade-off between inventory shortages and excess inventory is the essence of inventory management.

The inventory management problem has long been studied, even before 1960~\cite{scarf1958min,arrow1958studies}. Classically, this problem has been solved by operations research methods, such as dynamic programming~\cite{scarf1958min,DP_OPT1960} and mathematical programming~\cite{IPA1995,sQ_1998,bertsimas2006robust,chu2015simulation}.
After the year 2000, particularly from 2010, methods based on reinforcement learning~\cite{RL_1996,RL_1998} have also become mainstream for inventory management~\cite{giannoccaro2002inventory,jiang2009case,kemmer2018reinforcement,sultana2020reinforcement}. All the above methods treat the demand for products as time-series data, and the reinforcement learning agent learns generic delivery strategies for products and retailers.

The main advantage of the reinforcement learning approach is its adaptivity. The learned policy can be applied to unknown products and retailers. In reality, it is common for unknown products and retailers to emerge.  Manufacturers develop new products or discontinue old ones, and new retailers are opened or existing ones are closed.
Learning a generic strategy is more difficult than learning an individual strategy. However, for a system with a large number of products and retailers, learning generic strategies is much faster than learning multiple individual strategies. 
Another advantage of reinforcement learning is its flexibility since it can dynamically determine the number of products to order from the factory. Therefore, compared to most classical methods, it can easily respond to fluctuations in demand.

Regardless of the success of reinforcement learning in systems with a large number of products, no known algorithm can be used when both the number of products and the number of retailers are large. This is an important challenge that should be dealt with in the management of real warehouses. Therefore, our objective is to give an efficient and high-quality inventory management algorithm for such a huge and realistic system.

\subsection{Our Result}

We propose an optimization algorithm for warehouse inventory management that can be used in cases where both the number of product types $P$ and retailers $R$ are large.
The input to the algorithm is time-series data representing the demand history of products from a retailer. At each time step, the algorithm outputs the quantities of the products ordered from a factory. Because the demand history is given sequentially, the algorithm cannot see future demand quantities, although it can know the predicted values.
It should be noted that this study concentrates on optimizing warehouse inventory management and not retailer inventory management. The reason for this is that manufacturers and retailers are often different companies, and manufacturers cannot control retailers' strategies.

Our algorithm is based on reinforcement learning. To deal with multiple products, we construct a single learning agent that manages an inventory of all products. To do this, we consider both demand data and forecasts from retailers as well as information about each product (price, inventory cost, etc.) as observations during the learning process. In this manner, the learned delivery strategy can be applied to the inventory management of any product.

The problem with a large number of retailers is that the computational complexity of the simulation at each step of the learning process is large. At each time step, we need to simulate the delivery plan for all products and all retailers. This simulation takes $O(PR)$ time. In a huge system, this overhead cannot be ignored.

To resolve this problem, we reduced the computational complexity by approximating products other than the ones we were simulating. By doing this, the computational complexity of the simulation at each step was reduced to $O(R)$ time.

For the training algorithm, we used the PPO algorithm~\cite{PPO_2017}. For the training and evaluation data, we used the data of 100 products and 100 retailers, respectively. The training and evaluation data were independent of each other. We used real and artificial data. Our real data were generated from the sales data of Instacart~\cite{Kaggledata}, which is an online shopping service. Our artificial data focused on seasonal fluctuation of the demand. 

As a baseline, we used an algorithm based on base-stock policy. Intuitively, this method aims to maintain a constant amount of inventory at all times. To determine the target amount of inventory for all products, we used parameter optimization with Optuna~\cite{Optuna_2019}.
Our algorithm showed better performance than the baseline, which indicates that our learning method with retailer approximation is useful to handle large supply chain networks.

\subsection{Related Work}

\paragraph{Inventory Management}

Classically, operations research approaches are mainstream solutions for the inventory management problem and have been successful for relatively small problems.
In early studies focused on the inventory management problem, Scarf~\cite{scarf1958min} proposed a dynamic programming algorithm to solve the inventory management problem. Thereafter, Clark and Scarf~\cite{DP_OPT1960} proved the optimality of the algorithm.
It is noteworthy that their work deals with conditions where there is no uncertainty in the input.

Rule-based approaches have also succeeded in operation research-based classical inventory management. This kind of approach fixes an inventory management policy and optimizes its parameters. Glasserman and Tayur~\cite{IPA1995} used \emph{infinitesimal perturbation analysis}~\cite{IPA_1988} to optimize the target inventory amount, called \emph{base-stock value}. Such continuous optimization methods have been extensively studied for a variety of policies, models, and constraints~\cite{sQ_1998,bertsimas2006robust,chu2015simulation}.

 
The learning-based approach for inventory management has been studied since the year 2000. In this approach, the demand data is treated as a time series, and the order quantity at each time step is determined using reinforcement learning~\cite{RL_1996,RL_1998}. To our knowledge, Pontrandolfo et al.~\cite{RLsup_2002} first applied reinforcement learning to the inventory management problem. Jiang and Sheng~\cite{jiang2009case} proposed a method combining reinforcement learning and classical $(s,Q)$-policy~\cite{sQ_1998}. Mortazavi et al.~\cite{mortazavi2015designing} applied reinforcement learning to inventory management on a four-echelon supply chain network. Sultana et al.~\cite{sultana2020reinforcement} designed an algorithm that works fast and effectively even with multiple products by learning a generic strategy for each product; they also verified experimentally that their algorithm gives good results on real data. 
However, no previous research considers the supply chain network that both the number of retailers and products are large.



\subsection{Organization}

The remainder of this paper is organized as follows. First, we introduce our model of the supply chain network in the case of multi-product and multi-retailers. In the next section, we present our proposed algorithm. Thereafter, we present our experimental methods and results on real data. Finally, we conclude this paper by giving possible future research directions.

\section{The Model}

In this section, we explain how we model a supply chain network.

\subsection{Overview}

Our supply chain network is modeled using $P$ products, a warehouse, and $R$ retailers. Each product is produced in the corresponding factory and delivered to the warehouse. The warehouse then delivers the products to all the retailers.
The warehouse and the retailers manage the inventory of all $P$ products. The goal of the inventory management algorithm is to control the warehouse inventory by determining the quantity of products the warehouse will order from the factory and the quantity it will ship to the retailers. We express the quantity of the products in kilograms.

The quantity of demand at the retailers is given as time series data over $T$ days. In determining the action to be taken on the day $t$, the inventory management algorithm can consider the quantity demanded up to day $t-1$. The algorithm cannot know the quantity to be demanded in the future, although it may predict it.

We assign an agent to each retailer, and we call these agents \textit{retail agents}. Once a day, each retail agent decides how much to order from the warehouse for all products.
For each product, we also assign an agent that manages the quantity of inventory for that product in the warehouse. We call these agents \textit{product agents}.
It is noteworthy that this does not model a situation with multiple physical warehouses.
There is only one physical warehouse, although there are as many agents managing it as products.

Considering the decisions of the retail agents, each product agent decides once each day how much of the corresponding product to ship to each retailer.  To maintain inventory, it also decides the amount of the corresponding product to order from the factory. The products shipped by the warehouse to the retailers will arrive at the retailers after a certain number of days. This depends on the retailer since it represents the time it takes for a truck to go from the warehouse to the retailer. Similarly, products ordered by the warehouse from the factory will arrive at the warehouse after certain number of days. This depends on the product since it represents the time it takes for the factory to receive the order, produce the product, and deliver it to the warehouse.

The \emph{gain} of each product agent per day is determined by two factors. The first is the quantity of products shipped to the retailers. The \emph{profit} made by the warehouse for that day is formulated as the quantity multiplied by the selling price per kilogram. The other is the \emph{inventory cost}, which is proportional to the quantity of the inventory remaining in the warehouse that day. The gain of the product agent for each day is the profit minus the inventory cost. The total gain of the product agent is the sum of the gains over $T$ days, and the gain of the warehouse is the sum of the total gains of all the product agents.

Our goal is only to design an inventory management algorithm for the product agent, which means the inventory management of the warehouse and not the retail agent.
This is because, in many real-world situations, the warehouse does not have the authority to control the retailer's inventory management strategy. This is a common situation when the warehouse is controlled by the manufacturer and the retailers (e.g., convenience stores) are separate companies.

Therefore, we do not concentrate on optimizing the behavior of retail agents. Instead, we fix the behavioral strategies of the retail agents. We set parameters that are specific to each retail agent, however they are not meant to be used by product agents to define the strategy of each retail agent; rather, those parameters are for each retail agent to determine its actual behavior from a fixed behavioral strategy.

\subsection{Product Agent}

For each product $k\in \{1,\dots, P\}$, we have a product agent $k$ corresponding to that product. Each product agent $k$ has three values $price_k, stockcost_k\in \mathbb{R}_{\geq 0}$ and $delay_k\in \mathbb{Z}_{>0}$. 
These values represent information about the product $k$. Specifically, $price_k$ represents the selling price of the product $k$ per kilogram. $stockcost_k$ represents the stock cost per kilogram of stock of the product $k$ left in the warehouse at the end of each day. $delay_k$ represents the number of days from the day the product agent $k$ orders the product $k$ from the factory to the day it arrives at the warehouse. We assume that the products can be shipped in any portion -- that is, the quantity of products that are ordered or shipped can be any non-negative real number, as long as there is enough inventory.

The behavior of the product agent $k$ on each day $t$ is as follows. First, the agent receives the products ordered from the factory on the day $t-delay_k$. Based on this information, the product agent $k$ determines the quantity $order_{t,k}\in \mathbb{R}_{\geq 0}$ of product $k$ to be ordered from the factory at day $t$. We assume that the factory has an infinite production capacity (the value $order_{t,k}$ can take any non-negative real value). Thereafter, the agent checks the quantity of product $k$ ordered by the retail agents at day $t$. Then, for each retailer $i\in\{1,\dots, R\}$, the agent determines the quantity $ship_{t,k,i}\in\mathbb{R}_{\geq 0}$ of the product $k$ to be shipped to retailer $i$ at day $t$.
If we denote the quantity of the inventory of products $k$ remaining in the warehouse at day $t$ by $stock_{t,k}$, the updated formula for $stock_{t,k}$ on day $t$ becomes

\begin{align*}
    stock^+_{t,k} &= stock_{t-1,k} + order_{t-delay_k,k}\\
    stock_{t,k} &= stock^+_{t,k} - \sum_{i\in \{1,\dots, R\}}ship_{t,k,i}.
\end{align*}
Using these values, the gain of the product agent $k$ at day $t$ can be written as
\begin{align}
    price_k\times \sum_{i\in \{1,\dots, R\}}ship_{t,k,i} - stockcost_k\times stock_{t,k},\label{eq:prod_reward}
\end{align}
where the first and second terms represent the profit and inventory cost, respectively.

To concentrate on managing the quantity of products that the product agent orders from the factory, we fix a strategy to ship products to the retailers. Intuitively, our strategy is to ``fulfill orders as much as possible.'' This is the natural strategy for maximizing the profit of the product agent because the price of a product per kilogram does not change daily, and leaving excess inventory will only increase the cost of inventory.

Let us formalize this idea. When the retail agents order the product $k$ from the warehouse, the product agent $k$ we design will ship the product $k$ as long as the warehouse has the product $k$ in stock; that is, if we denote the quantity of the product $k$ ordered by retail agent $i$ at day $t$ by $request_{t,k,i}$, we have

\begin{align*}
    ship_{t,k} &:=\sum_{i\in \{1,\dots, R\}}ship_{t,k,i} \\
    &= \min\left(
        stock^+_{t,k},\sum_{i \in \{1,\dots, R\}} request_{t,k,i}\right).
\end{align*}

When $ship_{t,k}=\sum_{i\in \{1,\dots, R\}}request_{t,k,i}$, it means that on day $t$, the product agent $k$ has enough inventory to fulfill all the retail agents' orders. In that case, the product agent $k$ ships exactly $request_{t,k,i}$ quantity of products to each retail agent $i$.
In contrast, if $ship_{t,k}<\sum_{i\in \{1,\dots, R\}}request_{t,k,i}$, it means that the inventory held by the product agent $k$ on day $t$ is insufficient to fulfill the orders of all the retail agents. In that case, the product agent $k$ we design will ship all of its inventory on the day $t$ in such a manner that the quantity of shipments is proportional to the quantity of orders from each retail agent. Formally, the value $ship_{t,k,i}$ is represented as

\begin{align*}
    \min\left(1, \frac{stock^+_{t,k}}{\sum_{i\in \{1,\dots, R\}}request_{t,k,i}}\right) \times request_{t,k,i}.
\end{align*}

To help the product agent perform better inventory management, we give the product agent a prediction of the quantity of products that will be ordered by the retailer. Specifically, for the product agent $k$ on day $t$, this prediction is given as follows. Let $predictdays\in \mathbb{Z}_{>0}$ be a constant. For all integers $t'$ such that $t\leq t'<predictdays$, the prediction of the value of
\begin{align*}
    \sum_{i\in \{1,\dots, R\}}request_{t',k,i}
\end{align*}
is given by $predict_{t,t',k}$.

We should mention that this prediction may not always be accurate. On the contrary, we do not require any restrictions on this prediction. The predicted quantity of products ordered by retailers, $predict_{t,t',k}$, may be exactly the same as the quantity actually ordered, $\sum_{i\in \{1,\dots, R\}}request_{t',k,i}$. It may also happen that there is no correlation with the actual quantity ordered (for example, $predict_{t,t',k} = 0$ holds for all $t$, $t'$ and $k$). All variables are summarized in Table~\ref{prod_var_table}.

\begin{table}[hbtp]
    \caption{Variable summary of product agent}\label{prod_var_table}
    \centering
    \begin{tabular}{|l|l|}\hline
        $price_k$ & Price of $k$ per kilogram \\ \hline
        $stockcost_k$ & Inventory cost of $k$ per kilogram \\ \hline
        $delay_k$ & \#Days required for delivery from factory \\ \hline
        $stock_{t,k}$ & Inventory quantity of $k$ on day $t$\\ \hline
        $order_{t,k}$ & Order quantity of $k$ from factory on day $t$\\ \hline
        $ship_{t,k,i}$ & Quantity of $k$ shipped to retailer $i$ on day $t$\\ \hline
        $request_{t,k,i}$ & Quantity of $k$ requested from retailer $i$ on day $t$\\ \hline
        $predict_{t,t',k}$ & Prediction of $\sum_{i\in \{1,\dots, R\}request_{t',k,i}}$ on day $t$\\ \hline
        $predictdays$ & Period order prediction is available\\ \hline
    \end{tabular}
\end{table}

\subsection{Retail Agent}

For each retailer $i\in\{1,\dots, R\}$, we have a retail agent $i$ corresponding to that retailer. Each retail agent $i$ has $2$ values $trucksize_i\in \mathbb{R}_{\geq 0}$ and $delay_i\in \mathbb{Z}_{>0}$.
These values represent information about the retailer $i$. Specifically, $trucksize_i$ represents the maximum quantity of products that can be loaded onto a truck from the warehouse to the retailer $i$. A truck can ship any combination of any number of product types in any non-negative real quantity, as long as the total weight does not exceed $trucksize_i$ kilograms.
The $delay_i$ represents the number of days between the day the retailer orders the products from the warehouse and the day the products arrive at the retailer $i$.

The actions of the retail agent $i$ on each day $t$ are as follows. First, for each product $k$, the agent receives the product that the warehouse shipped to the retailer $i$ on the day $t-delay_i$. It is noteworthy that the quantity of products that the retail agent receives on the day $t$ may be different from the quantity of products that it ordered from the warehouse on the day $t-delay_i$. It should also be noted that this is a problem specific to the retail agent; this problem does not occur for product agents since the factory has an unlimited inventory.

Therefore, for all products $k$, the retail agent $i$ determines the quantity $request_{t,k,i}$ of products $k$ to be ordered by the retailer $i$ to the warehouse on day $t$. Thereafter, the retail agent sells the products to the customers. We denote the quantity of product $k$ sold by retailer $i$ to customers on day $t$ by $sales_{t,k,i}$.
Let $stock_{t,k,i}$ be the quantity of product $k$ in stock of retailer $i$ on day $t$, then the update formula of $stock_{t,k,i}$ on day $t$ is represented as

\begin{align*}
    stock^+_{t,k,i} &= stock_{t-1,k,i} + ship_{t-delay_i,k,i}\\
    stock_{t,i,k} &= stock^+_{t,k,i} - sales_{t,k,i}.
\end{align*}

We fix a strategy for the retail agents to sell products to customers. Intuitively, our strategy is to ``satisfy the demand as much as possible.'' Formally, if $demand_{t,k,i}$ is the quantity of the demand from customers for product $k$ at retailer $i$ on day $t$, then $sales_{t,k,i}$ is represented as
\begin{align*}
    \min\left(demand_{t,k,i}, stock^+_{t,k,i}\right).
\end{align*}
We emphasize that, in this scenario, the product agents cannot control the behavior of the retail agents; therefore, we do not optimize the behavior of the retail agents but rather fix their strategies. For the same reason, we do not strictly define the gains of the retail agents.
Before describing the strategies of the retail agents, we must first describe the scenario for the retailers.
It is important to note that this description is not necessary to understand the mathematical detail of our model. Rather, this description of the scenario helps us understand the validity of the retail agent's strategy.

In this scenario, a retail agent's \emph{gain} is determined by the \emph{profit} made from selling the product to the customer, the \emph{inventory cost}, and the \emph{shipping cost} from the warehouse.
First, the store's profit is determined by the quantity of products sold to customers. Specifically, the profit of the retail agent $i$ is written as a non-negative combination of $sales_{t,k,i}$ over each day $t$ and each product $k$.
Similarly, the inventory cost of the retail agent $i$ is written as the non-negative combination of $stock_{t,k,i}$, the quantity of unsold products at the end of each day, over each day $t$ and each product $k$.

Finally, we design the \emph{shipping cost} from the warehouse, which is the cost of moving a truck from the warehouse to the retail store to deliver the products. This cost depends only on whether the quantity of products to be shipped that day is $0$ or not. For a day $t$ and a retailer $i$, if there is a product $k$ with $ship_{t,k,i}>0$, then there is a constant cost for the retailer $i$ on day $t$.
This captures the following idea: if the total quantity of shipped products is the same, it is more efficient to ship much of them at once than in small portions. This term is particularly important for small retailers, where the quantity of products shipped from the warehouses is relatively small, and therefore it is impossible to fill a truck with products demanded in a day.

Now, let us introduce the strategy of the retail agent we have designed. Each retail agent $i$ has a real value $basestock_{k,i}\in \mathbb{R}_{>0}$ for each product $k$. We set $stock_{0,k,i}=basestock_{k,i}$ for all product $k$ and retailer $i$. As in the classical base-stock policy~\cite{hopp2011factory}, the retail agent attempts to behave in manner such that the quantity of inventory, including the unarrived portion of each product, is kept at a constant value. Specifically, the quantity of product $k$ that retail agent $i$ intends to order from product agent $k$ on day $t$ is denoted by
\begin{align*}
    &lack_{t,k,i} := \\
    &\max\left(0, basestock_{k,i} - stock_{t,k,i} - \sum_{t'=t-delay_i}^{t-1}ship_{t,k,i}\right).
\end{align*}

If we follow this strategy, however, we will frequently have to ship only a small quantity of products at one time. It can be observed that every time a retailer sells a non-zero quantity for a product, a shipment happens. To avoid this problem, we modify the strategy slightly so that an order is placed only when the total quantity of products that retailer $i$ intends to order on the day $t$ exceeds the capacity of the truck. Formally, the retail agent $i$ will place an order only when 

\begin{align*}
    lack_{t,i} := \sum_{k\in \{1,\dots, P\}}lack_{t,k,i}\geq trucksize_i.
\end{align*}
Furthermore, the quantity of products to be ordered at that time is normalized so that it fits exactly into the capacity of the truck. Specifically, when the above condition is satisfied, the quantity $order_{t,k,i}$ of product $k$ that the retail agent $i$ orders from the warehouse on the day $t$ is

\begin{align*}
    \frac{trucksize_i}{lack_{t,i}}\times lack_{t,k,i}.
\end{align*}
By this construction, if the warehouse always has enough inventory, then the trucks shipping products from the warehouse to the retail stores will always be full. All variables are summarized in Table~\ref{retail_var_table}.

\begin{table}[hbtp]
    \caption{Variables Summary of Retail Agent}\label{retail_var_table}
    \centering
    \begin{tabular}{|l|l|}\hline
        $trucksize_i$ & Capacity of truck used by $i$ \\ \hline
        $delay_i$ & \#Days required for delivery from warehouse\\ \hline
        $stock_{t,k,i}$ & Inventory quantity of $k$ on $i$ on day $t$\\ \hline
        $sales_{t,k,i}$ & Sales quantity of $k$ on $i$ on day $t$\\ \hline
        $demand_{t,k,i}$ & Demanded quantity of $k$ on $i$ on day $t$\\ \hline
        $basestock_{k,i}$ & Base-stock value of $k$ on $i$ on day $t$\\ \hline
    \end{tabular}
\end{table}

\section{Methodology}

In this section, we describe our reinforcement learning-based algorithm for determining the strategies of product agents. Our goal is to learn an inventory management strategy that can be used universally by all product agents.

\subsection{General Algorithm}

Let $k$ be a product. First, we consider the inventory management of a product agent $k$ as a Markov decision process by fixing the actions of all product agents except $k$.
The state space is defined as the combination of all the variables introduced in the previous section.
The action space is the quantity $order_{t,k}\in \mathbb{R}_{\geq 0}$ that the product agent $k$ orders from the warehouse on day $t$.
The transition function is as described in the previous section.
The reward function is same as the gain of the product agent $k$ in the previous section.

In our algorithm, a single reinforcement learning agent learns using a Markov decision process for all product agents $k$. Specifically, our reinforcement learning agent repeats the following: The agent randomly chooses $k\in\{1,\dots, P\}$ and updates the policy by simulating the observations and actions of the product agent $k$ for $T$ days. The policy obtained in this manner is expected to be a high-quality policy for all product agents $k$.

It is noteworthy that our goal was to design an inventory management algorithm applicable to a large number of products and retailers. In such a case, the computational complexity of simulating the entire system would be large. Specifically, to simulate the transition of the system from day $t$ to day $t+1$, it would be necessary to calculate $request_{t,k,i}$ for all products $k$ and all retailers $i$. Naively, this would take $O(PR)$ time because of the following reason: the quantity $request_{t,k,i}$ of product $k$ of each retail agent $i$ is affected by the inventory quantity of other products of that retail agent, which depends on the inventory amount of the other products on the warehouse. Consequently, it depends on the amount of inventory of all products of all retail agents. When the numbers of products $P$ and retailers $R$ are both large, this becomes a bottleneck for reinforcement learning.

We considered reducing the time complexity. Before starting the training, we first simulated the behavior of all retail agents. Then, for all days $t$, $k$ and $i$, we recorded the quantity $request^*_{t,k,i}$ of the product $k$ ordered from the warehouse by retailer $i$ on day $t$. In this simulation, we assumed that the product agents have infinite inventory. It is worth remarking that this simulation was executed only in the training phase and not in the evaluation phase.

In the training phase, we used $request^*_{t,k,i}$ to compute an approximation of the retail agent's behavior. Intuitively, our approximation is given by fixing the list of days to move truck. For any $t$, $k$ and $i$, an approximation of $request_{t,k,i}$ is given by
\begin{align*}
    request^a_{t,k,i} := \left(\begin{array}{cc}
        lack_{t,k,i} & \left(\sum_{k\in \{1,\dots, P\}}request^*_{t,k,i} > 0\right) \\
        0 & (\text{otherwise})
    \end{array}\right..
\end{align*}
When performing simulations for the product agent $k$, for any $t$ and $i$, $request^a_{t,k,i}$ can be computed in constant time by remembering whether $\sum_{k\in \{1,\dots, P\}}request^*_{t,k,i}$ is zero or not. This is in contrast to the fact that we needed to simulate all other product agents to obtain the exact value of $request_{t,k,i}$.
During the training of the product agents, we used the approximation $request^a_{t,k,i}$ instead of $request_{t,k,i}$ for the quantity of orders placed by retailer $i$ to warehouse $k$ at time $t$. Using this approach, the computational complexity in simulating the transition of the system from day $t$ to day $t+1$ was reduced from $O(PR)$ to $O(R)$.

We had to justify this approximation by evaluating our algorithm on a different set of data from the training data. In the evaluation phase, we used the exact value of $request_{t,k,i}$ as the quantity of orders placed by retailer $i$ to warehouse $k$ at time $t$. We observed that the computation time of $request_{t,k,i}$ in the evaluation phase did not matter because the evaluation phase runs only $1$ time of the simulation of the whole system for $T$ days.

\paragraph{Observation}

Let $k$ be the product agent currently being simulated by the training algorithm and $t$ the day being simulated. The observation given to our reinforcement learning agent consists of three parts. The first is the information about the product agent $k$, consists of $price_k$, $stockcost_k$ and $delay_k$.
It should be noted that the subscript $k$ of the product agent itself is not included in the observation to create a learning agent that can address unknown products.
The second is the quantity of inventory of the products $k$ currently held by the product agent $k$, including those that have been ordered from the factory, but not yet received. This quantity is represented as 
\begin{align*}
    stock_{t,k} + \sum_{t' = t - delay_k}^{t-1}order_{t',k}.
\end{align*}
The third is the order prediction, as explained in the section on the product agent model.
These observations are summarized in Table~\ref{obs_table}.





\begin{table}[hbtp]
    \caption{Observation for warehouse $k$ on day $t$}\label{obs_table}
    \centering
    \begin{tabular}{|l|c|}\hline
        product information & $price_k, stockcost_k, delay_k$ \\ \hline
        current stock & $stock_{t,k} + \sum_{t' = t - delay_k}^{t-1}order_{t',k}$ \\ \hline
        order prediction & $predict_{t, t',k}$\\ 
         & $(t < t'\leq t + preeictdays)$\\ \hline
    \end{tabular}
\end{table}

\paragraph{Action and Reward}

Let $k$ be the product agent that the learning agent is currently simulating and $t$ be the day. The action of the learning agent is the quantity $order_{t,k}$ of product $k$ to be ordered from the factory on the day $t$. To improve the efficiency of learning, we discretize this action; we prepare the constant $maxorder\in \mathbb{R}_{>0}$ and choose the parameter $x\in \{0, 1\}$. Here, the value of $order_{t,k}$ is equal to $maxorder$ if $x=1$ and $0$ otherwise.
Furthermore, under the same settings, the reward given to the learning agent is identical to the reward for the product agent $k$ on the day $t$, given in~(\ref{eq:prod_reward}).

\section{Experiment}

In this section, we describe the experimental methods and results.

\subsection{Data Preparation}

\paragraph{Real Data}

The real data used for training and evaluation were based on the sales data of Instacart~\cite{Kaggledata}, an American mail-order service specializing in food delivery, as in Sultana et al.~\cite{sultana2020reinforcement}.
For this study, the original data used was the order history of products for Instacart customers. Each order was given as a tuple of the index of the customer, including the set of products ordered and the date the order was placed. The date the order was placed was not given directly, but as a pair of the number of days that had elapsed since the customer placed the first order during the period covered by the data, and the day of the week.

To process the data, the customers whose data were to be considered were first narrowed down. Only the data from customers whose orders had been placed for a sufficiently long period of time, specifically more than $350$ days, were used. Then, the order period was unified across the different customers, assuming that each customer placed their first order in the first week of the period considered for the processed data.
The reason for this setting is that, in the original data, the period between the first and the last order for any customer was within $365$ days. Therefore, although unstated, we assumed that this data represented orders for a particular year. Because we wanted to deal with seasonal variations in the quantity of orders for each product, we only want to use customers who can be somewhat certain of the day of their first order. Specifically, under our assumption, the customers we use place their first order within the first $15$ days of the year (in original data). This way, the seasonal information would not be lost during preprocessing.

Each of our retailers corresponded to several customers. Specifically, if a retailer $i$ corresponds to a set of customers $X$, then for each product $k$, the value of $demand_{t,k,i}$ is the sum of the number of products $k$ ordered by customers belonging to $X$ on the day $t$.

Next, the products were narrowed down. Only products that appeared in the original data neither too few nor too many times were used. 
This was done for the following reasons.
First, since real-world retailers do not sell products that customers order a few times, it is meaningless to manage the inventory of such products.
In addition, we do not expect our algorithm to manage the inventory of products that appear in an extremely large number of orders. Because the number of such products is small and their impact on the gain is large for retailers, it is better to run a separate inventory management algorithm instead of applying a general-purpose algorithm.

We also thinned out our products further using products with large seasonal fluctuations in terms of order quantity. For each product, we computed a list of the total demand for all retailers for all consecutive $70$ days and kept only those products with relatively large standard deviations divided by mean. This was done in order to concentrate on inventory management for products with large seasonal demand fluctuations.
In this manner, we obtained the demand history data for $200$ products and $200$ retailers.

Since we have assumed that each customer's first order was placed during the first week, we cut off the demand data for the first week, which could be highly biased.
We also set $T=300$ and removed the demand data from the last period from the simulation. This was to generate future order prediction data to be given as observations to the learning agents even on the day $T$.

Finally, we scaled the retail demand data for each product so that the total demand for the product during the period given by the input was constant regardless of the product. The purpose of this operation was to allow the learning agent to handle observations, actions, and rewards for different products at the same scale. Remark that when dealing with unknown products, we need to be able to predict the total demand for the product in order to perform this operation. However, this is a reasonable assumption since we are assuming that we can predict the quantity of products requested from retailers.

It is noteworthy that the original data did not contain individual information for each product, such as price and weight. Therefore, we randomly set the information $price_k$, $stockcost_k$ and $delay_k$ for each product $k$. Considering that we artificially constructed the retailers, we also randomly set the information $trucksize_i$ and $delay_i$ for each retailer $i$.

\paragraph{Artificial Data}

We also used artificial data in our experiments, considering seasonal fluctuation in demand. Specifically, for all days $t$, products $k$, and retailers $i$, we set
\begin{align*}
    demand_{t,k,i} = \left(1+\cos\left(\frac{2\pi(t+off_{k})}{365}\right)\times fluc_{k,i}\right)\times C,
\end{align*}
where the $off_k$ is a parameter that determines the day of demand peak and is sampled uniformly from $[0,365]$, the $fluc_{k,i}$ is a parameter that determines the amount of demand fluctuation and is sampled uniformly from $[0,1]$, and $C$ is a constant.

\subsection{Training and Evaluation}

We split the data we created in the previous section into two sets and generated two sets of demand data for 100 products and 100 retailers (one for training and the other for evaluation). We used the PPO algorithm~\cite{PPO_2017} to train our learning agent for $1.5\times 10^7$ days, or equivalently $5\times 10^4$ episodes, using the training data.
In the PPO algorithm we used MLP with 100 units and two hidden layers. As mentioned earlier, to speed up the simulation, we approximated the behavior of the retailers during the training. The performance of the resulting learning agent was then evaluated using the evaluation data and exact retailers simulation.

\subsection{Order Prediction}

The predicted order $predict_{t,t',k}$ of product $k$ for day $t' > t$, given to product agent $k$ on each day $t$, is given by the sum $\sum_{i\in \{1,\dots, R\}}demand_{t',k,i}$ of the demands for product $k$ in all retailers on day $t'$
This is a simplification of the situation so that we can concentrate on evaluating the warehouse inventory management strategy itself, rather than on evaluating the accuracy of the order prediction. It may seem as though our algorithm uses future demand data as an observation. However, it should be noted that $predict_{t,t',k}$ is different from the actual order quantity $require_{t',k}$ to the product agent $k$; this is consistent with the ordering strategy for retailers.

\subsection{Baselines}

In this section, we describe our baseline algorithms.

\paragraph{Base-stock Policy}

We used an algorithm based on the base-stock policy as a baseline for the behavior of product agents. This policy is simple but cannot deal with seasonal demand fluctuations. Each retail agent $k$ has a base-stock quantity $basestock_k$. On each day $t$, the retail agent $k$ orders 
\begin{align*}
    \max\left(0,basestock_k - stock_{t,k} - \sum_{t' = t-delay_k}^{t-1} order_{t',k}\right)
\end{align*}
quantity of products so that the quantity of stock, including the quantity ordered from the factory but not yet arrived, is $basestock_k$.

In a system with only one type of product, various algorithms are known to determine the value of $basestock_k$. However, we are considering a system that is too large for the separate optimization of the behavior of each product agent. Instead, we fix the form of the formula for determining the value of $basestock_k$ from the information about product $k$ and optimize the parameters in the formula. Specifically, this can be expressed as follows:
\begin{align*}
    basestock_k = \frac{\sum_{t=1}^{T}\sum_{i\in \{1,\dots, R\}}request^*_{t,k,i}}{T} \times x \times delay_k,
\end{align*}
where $request^*_{t,k,i}$ is a quantity of product $k$ ordered by retailer $i$ at time $t$, assuming that the warehouse has infinite inventory. 

This is based on the following intuition. Consider the case where the total quantity of requests placed by retailers for product $k$, $\sum_{i\in \{1,\dots, R\}}request_{t,k,i}$, is constant regardless of day $t$. The optimal strategy of the product agent $k$ is to order that quantity on each day $t$. This is exactly the strategy obtained by setting $x=1$ in the above formula.
Then, we optimize the parameter $x$ using Optuna~\cite{Optuna_2019}.

\paragraph{Oracle Strategy}

For another comparison, we used the case where each product agent knows all future requests from the retail agents. In this case, the optimal strategy for the product agent $k$ is to have no inventory at all and to ensure that on each day $t$, exactly the same quantity of product $k$ as requested from the retail agents arrives from the factory. Formally, we set $order_{t,k} = request_{t + delay_k, k}$
for all $t$.
Equivalently, this baseline can be viewed as the case where the product agents have infinite inventory and zero inventory cost.
It is noteworthy that this baseline is an optimal strategy in situations where we cannot control the retail agents' strategy. Therefore, we do not aim to find a better strategy than this baseline.

Since we consider the supply chain network with many products, our environment is different from the Sultana et al.'s one~\cite{sultana2020reinforcement}. Therefore, we do not directly compare our result between their result.


\subsection{Results}
Table~\ref{tab:score_1} and table~\ref{tab:score_2} show the evaluation results for our learning agent, the base-stock algorithm-based agent, and the oracle agent on the real and artificial data, respectively. Each number in the table is the average of the sum of the gains, or equivalently, the profits minus the inventory cost, of all product agents, where the average is taken over the days. These results show that our learning agents learned better strategies than the base-stock policy-based agents. More specifically, our learning agent achieves a better gain by fulfilling more requests from retailers than the base-stock policy-based agent.

\begin{table}[hbtp]\caption{Gains, profits and inventory costs by our training agent and baseline agents on real data.}\label{tab:score_1}
    \begin{tabular}{|c||c|c|c|}\hline
         & Our Alg & Base-stock & Oracle\\ \hline \hline
        Av. Gain  & $\mathbf{23153}$ & $23049$ & $25382$\\ \hline
        Av. Profit & $24790$ & $24314$ & $25382$\\ \hline
        Av. Inventory Cost & $1637$ & $1264$ & $0$\\ \hline
    \end{tabular}
\end{table}

\begin{table}[hbtp]\caption{Gains, profits and inventory costs by our training agent and baseline agents on artificial data.}\label{tab:score_2}
    \begin{tabular}{|c||c|c|c|}\hline
         & Our Alg & Base-stock & Oracle\\ \hline \hline
        Av. Gain  & $\mathbf{23040}$ & $21113$ & $27413$\\ \hline
        Av. Profit & $26787$ & $24408$ & $27413$\\ \hline
        Av. Inventory Cost & $3746$ & $3294$ & $0$\\ \hline
    \end{tabular}
\end{table}

\begin{figure}[hbtp]
    \centering
    \includegraphics[width = 4cm]{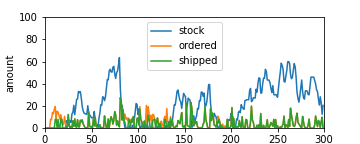}
    \includegraphics[width = 4cm]{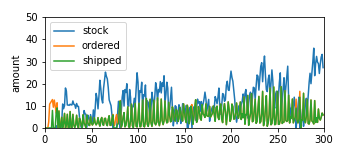}\\
    \includegraphics[width = 4cm]{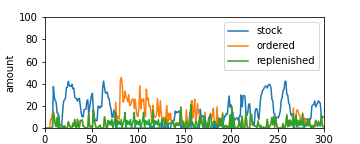}
    \includegraphics[width = 4cm]{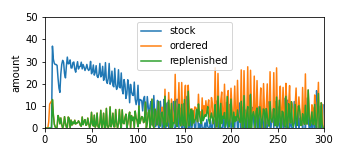}\\
    \includegraphics[width = 4cm]{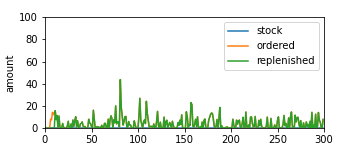}
    \includegraphics[width = 4cm]{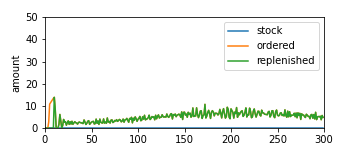}
    \caption{Inventory history of products. Left and right column represents a product from real data and artificial data, respectively. Figures in the first, second and the third row represents the result of our algorithm, base-stock policy-based algorithm and oracle algorithm, respectively.}
    \label{fig:result}
\end{figure}
Figure~\ref{fig:result} shows the inventory management history of some products when running our learning agent on the evaluation data. The horizontal axis represents the number of days, and the vertical axis represents the quantity of products. In the run results for product $k$, the blue line represents the current inventory $stock_{t,k}$, the orange line represents the total quantity of requests from retailers $\sum_{i\in \{1,\dots, R\}}request_{t,k,i}$, and the green line represents the total quantity of shipments to retailers. It is noteworthy that the overlap between the orange and green lines indicates that the request from the retailer has been completely fulfilled. By comparing the graphs in the first and second rows, you can see that our algorithm suppresses shortages of inventory due to seasonal fluctuation of demand compared to the base-stock policy.

\begin{figure}[hbtp]
    \centering
        \includegraphics[width = 8cm]{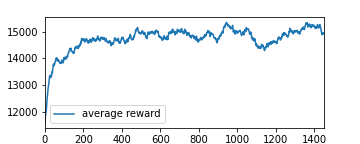}
        \caption{Learning Curve of real data. The horizontal axis represents the number of  learning steps (divided by $10^4$), and the vertical axis represents the mean of the learning agent's reward.}\label{fig:curve1}
    \centering
        \includegraphics[width = 8cm]{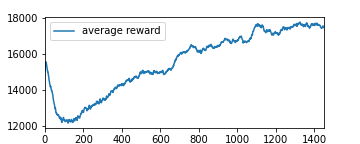}
        \caption{Learning Curves of artificial data. The horizontal axis represents the number of learning steps (divided by $10^4$), and the vertical axis represents the mean of the learning agent's reward.}\label{fig:curve2}
\end{figure}
During training, we evaluated the current agent's reward every 10,000 steps by using 10 random products agents. Figure~\ref{fig:curve1} and Figure~\ref{fig:curve2} shows the histories of the rewards. The horizontal axis represents the number of learning steps, and the vertical axis represents the mean of the learning agent's reward. For the sake of clarity, the mean of reward is shown as the average of 50 consecutive evaluations: this reduces the effect of sampling blur and produces a more continuous graph. It is noteworthy that this reward is different from the gain of the product agent in two means: First, this reward is evaluated not on the evaluation data, but on the training data. Second, this reward is an approximate reward, calculated by approximating retailers' behaviors.

\section{Conclusion}

In this paper, we proposed a warehouse inventory management algorithm. Our algorithm is based on reinforcement learning and can be applied to supply chain networks where the number of products and retailers are both large. We validated our algorithm on real and artificial data and demonstrated its superiority over the classical baseline.

There are three directions for future research. One is to run our algorithm on larger real data. Because of the size of the original real data, we only demonstrate the case of $100$ products and $100$ retailers. 

The second direction is to consider a hybrid of classical and learning-based methods. In our experiments, although the learning-based method was superior on average in our data, the superiority of the two methods varies from product to product. Therefore, if we have a good criterion to decide which products to apply the classical method and which products to apply the learning-based method, we will automatically have a better inventory management algorithm.

The third direction is when the constraints on the behavior between different product agents are not independent. For example, this would correspond to the case where there is a constraint on the inventory quantity in the entire warehouse. In this case, learning-based methods cannot learn by repeatedly simulating a single product agent. Hence, we have to fully simulate the system for a day, even during learning. For large systems, this is a difficult task, which makes inventory management of this type of large system a challenging problem.



\section{Acknowledgments}
We would like to thank Yasuhiro Fujita for advise on design of learning algorithms.

\bibliographystyle{ACM-Reference-Format} 
\bibliography{aaai22}

\end{document}